\documentclass[10pt,twocolumn,letterpaper]{article}

\usepackage[pagenumbers]{iccv} 

\definecolor{iccvblue}{rgb}{0.21,0.49,0.74}
\usepackage[pagebackref,breaklinks,colorlinks,citecolor=iccvblue]{hyperref}
\usepackage{multirow,adjustbox,diagbox,threeparttable}
\usepackage{amsfonts,dsfont,pifont,bm,bbm,mathrsfs,mathtools,nicefrac}

\definecolor{codegreen}{rgb}{0,0.6,0}
\definecolor{codegray}{rgb}{0.7,0.7,0.7}
\definecolor{codepurple}{rgb}{0.58,0,0.82}
\definecolor{backcolour}{rgb}{1.0,1.0,1.0}

\newcommand{\method}{{\sc{\textbf{Ross3D}}}\xspace}
\newcommand{\supp}{\textit{Supplementary Material}\xspace}
\definecolor{Light}{rgb}{0.99, 0.92, 0.95}
\renewcommand{\paragraph}[1]{\vspace{1.25mm}\noindent\textbf{#1}}

\definecolor{upcolor}{RGB}{57,182,74}
\newcommand{\up}[1]{\textcolor{upcolor}{$\uparrow$ #1}}
\newcommand{\down}[1]{\textcolor{red}{$\downarrow$ #1}}

\def\paperID{3069} 
\def\confName{ICCV}
\def\confYear{2025}

\title{
  \method: Reconstructive Visual Instruction Tuning with 3D-Awareness
}

\author{
Haochen Wang$^{1,2}$ \quad
Yucheng Zhao$^{3\dag}$ \quad
Tiancai Wang$^{3*}$ \quad
Haoqiang Fan$^3$ \\
Xiangyu Zhang$^{4,5}$ \quad
Zhaoxiang Zhang$^{1,2}$\thanks{Corresponding authors. $\dag$ Project lead.} \\[3pt]
$^1$NLPR, MAIS, CASIA \quad $^2$UCAS \quad $^3$Dexmal \quad $^4$MEGVII Technology \quad $^5$StepFun \\
{\small\texttt{
\{wanghaochen2022, zhaoxiang.zhang\}@ia.ac.cn
}} \quad {\small\texttt{wtc@dexmal.com}}\\[3pt]
Project Page: \small{\url{https://haochen-wang409.github.io/ross3d}}
}

\begin{document}
\maketitle
\begin{abstract}

The rapid development of Large Multimodal Models (LMMs) for 2D images and videos has spurred efforts to adapt these models for interpreting 3D scenes.
However, the absence of large-scale 3D vision-language datasets has posed a significant obstacle.
To address this issue, typical approaches focus on injecting 3D awareness into 2D LMMs by designing 3D input-level scene representations.
This work provides a new perspective.
We introduce \textbf{\underline{r}}ec\textbf{\underline{o}}nstructive vi\textbf{\underline{s}}ual in\textbf{\underline{s}}truction tuning with \textbf{\underline{3D}}-awareness (\method), which integrates 3D aware visual supervision into the training procedure.
Specifically, it incorporates cross-view and global-view reconstruction.
The former requires reconstructing masked views by aggregating overlapping information from other views.
The latter aims to aggregate information from all available views to recover Bird's-Eye-View images, contributing to a comprehensive overview of the entire scene.
Empirically, \method achieves state-of-the-art performance across various 3D scene understanding benchmarks.
%
%
More importantly, our semi-supervised experiments demonstrate significant potential in leveraging large amounts of unlabeled 3D vision-only data.
%
    
\end{abstract}
\section{Introduction}\label{sec:intro}

Embodied Artificial Intelligence systems are designed to effectively interact with physical environments~\cite{durante2024agent, liang2023code, singh2023progprompt, brohan2023rt, black2024pi_0}, offering transformative potential applications.
Central to these systems is the ability to understand 3D scenes comprehensively.
This involves both modeling spatial relationships between objects~\cite{chen2020scanrefer, huang2022multi, chen2021scan2cap} and comprehending the overall layout~\cite{azuma2022scanqa, ma2023sqa3d, yang2024thinking}, which is critical for enabling embodied agents to navigate, manipulate objects, and perform complex tasks in diverse environments.

\begin{figure}
    \centering
    \includegraphics[width=1\linewidth]{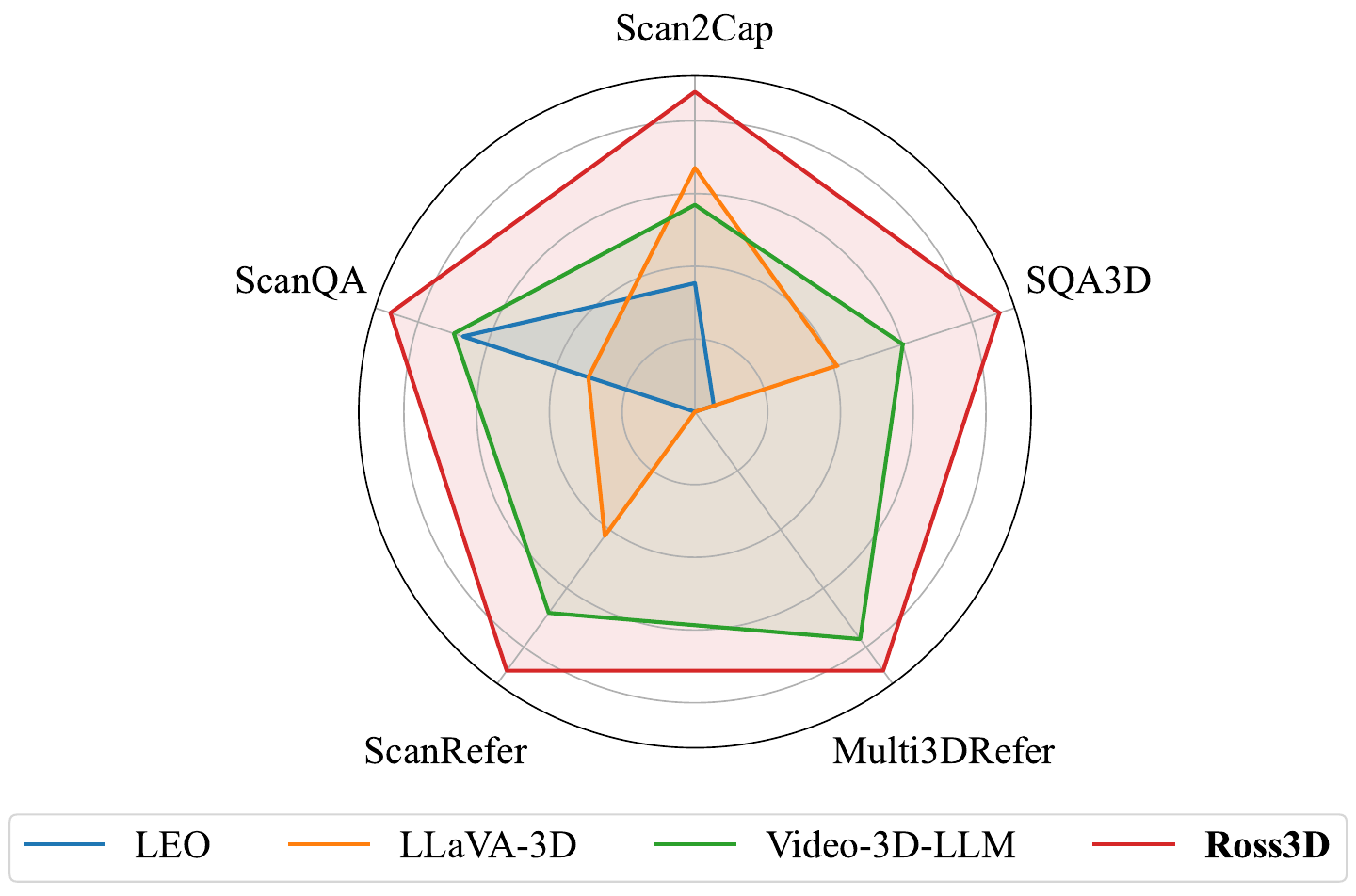}
    \vspace{-20pt}
    \caption{
    \textbf{Performance of \method} compared with state-of-the-art alternatives.
    We report EM on SQA3D~\cite{ma2023sqa3d}, CIDEr on ScanQA~\cite{azuma2022scanqa}, ROUGE on Scan2Cap~\cite{chen2021scan2cap}, Acc@0.25 on ScanRefer~\cite{chen2020scanrefer}, and F1@0.25 on Multi3DRefer~\cite{zhang2023multi3drefer}.
    With 3D-aware visual supervision, \method significantly outperforms other approaches across various benchmarks.
    }
    \label{fig:radar}
    \vspace{-10pt}
\end{figure}

\begin{figure*}[t]
    \centering
    \begin{subfigure}[b]{0.32\linewidth}
        \includegraphics[width=1\linewidth]{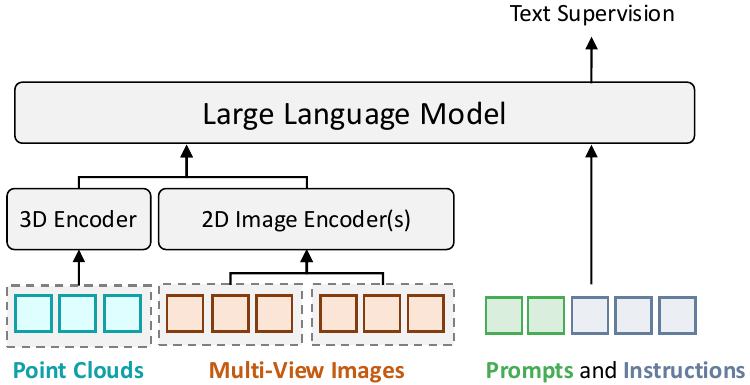}
        \caption{
        Point-based methods~\cite{zhang2024chatscene, hong20233d, fu2024scene} that fuse 3D point cloud features with 2D features.
        }
        \label{fig:3dllm}
    \end{subfigure}
    \hfill
    \begin{subfigure}[b]{0.32\linewidth}
        \includegraphics[width=1\linewidth]{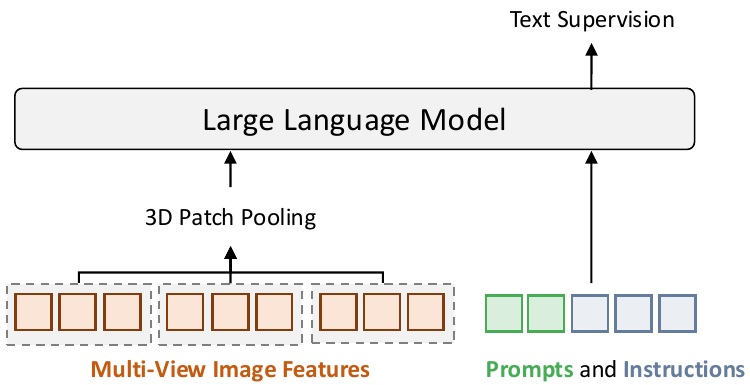}
        \caption{
        3D patch-based methods~\cite{zhu2024llava3d} that aggregate the 2D image features in the voxel space.
        }
        \label{fig:llava3d}
    \end{subfigure}
    \hfill
    \begin{subfigure}[b]{0.32\linewidth}
        \includegraphics[width=1\linewidth]{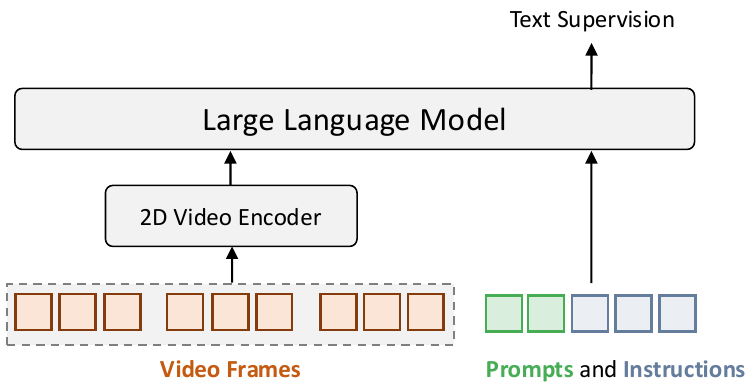}
        \caption{
        Video-based methods~\cite{zheng2024video3dllm, qi2025gpt4scene} that treat multi-view images as video sequences.
        }
        \label{fig:video3dllm}
    \end{subfigure}
    \begin{subfigure}[b]{1\linewidth}
        \includegraphics[width=1\linewidth]{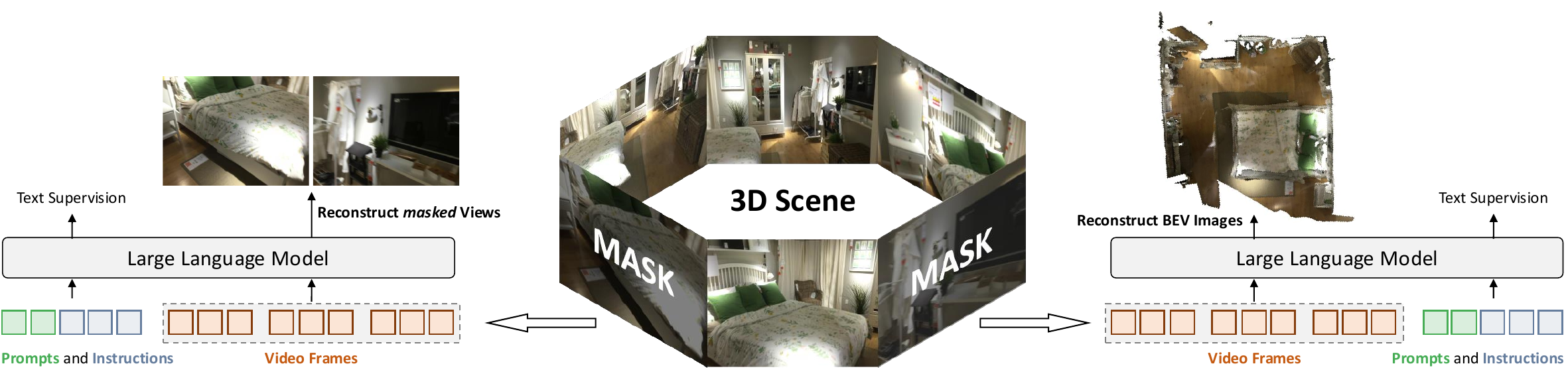}
        \vspace{-10pt}
        \caption{
        \textbf{The high-level idea of \method.}
        We inject vision-centric 3D-aware visual \textit{supervision signals} into 2D LMMs by incorporating 3D-aware pretext tasks, including cross-view (left) and global-view (right) reconstruction.
        }
        \label{fig:ours}
    \end{subfigure}
    \hfill
    \vspace{-18pt}
    \caption{
    \textbf{Conceptual comparison of our \method with popular paradigms.}
    Unlike previous methods that preliminarily focus on \textit{input-level} modifications to craft 3D-aware input representations, we incorporate 3D-aware visual pretext tasks.
    }
    \label{fig:comparison}
    \vspace{-10pt}
\end{figure*}

The remarkable success of Large Multimodal Modals (LMMs) in handling images~\cite{gpt4v, wang2024qwen2vl, chen2024internvl2.5, li2024llava, wang2025ross, wu2024deepseekvl2, gpt4o, claude, gemini} and videos~\cite{zhang2024llavavideo, zhang2025videollama, wang2024qwen2vl} has motivated researchers to adapt these models to interpret 3D scenes~\cite{zheng2024video3dllm, hong20233d, zhu2024llava3d, qi2025gpt4scene, zhang2024chatscene, fu2024scene}.
Similar to 2D LMMs, a straightforward approach is to develop 3D LMMs by projecting 3D point cloud features into the feature space of Large Language Models (LLMs) using point cloud-text pairs~\cite{chen2024ll3da, wang2023chat, huang2023leo, xu2024pointllm}.
However, unlike the abundance of large-scale 2D image-text pairs, 3D datasets remain extremely limited. 
Moreover, there are no powerful pre-trained 3D point cloud encoders, such as CLIP~\cite{radford2021learning} in 2D, to provide strong language-aligned 3D features, leading to unsatisfactory performance shown in \Cref{fig:radar}.
Therefore, researchers~\cite{zhu2024llava3d, zheng2024video3dllm, qi2025gpt4scene} begin to focus on building a 3D LMM based on the strong 2D priors from 2D LMMs.
In such a setting, incorporating 3D-awareness into LMMs originally trained on 2D data becomes a significant challenge.
To address this issue, previous attempts preliminary focus on \textit{crafting 3D-aware input representations} as illustrated in \Cref{fig:comparison}, including fusing 3D point cloud features with 2D image features~\cite{zhang2024chatscene, hong20233d, fu2024scene} in \Cref{fig:3dllm}, aggregating 2D features into 3D voxel spaces~\cite{zhu2024llava3d} with 3D patch pooling in \Cref{fig:llava3d}, and treating multi-view images as video sequences~\cite{zheng2024video3dllm, qi2025gpt4scene} in \Cref{fig:video3dllm}.


Despite these advancements, these kinds of input-level modifications \textit{alone} are insufficient to learn genuine 3D awareness.
This is because the inherent inductive bias toward 2D data in LMMs impedes the effective integration of 3D information.
As a result, even with enhanced inputs, LMMs struggle to produce optimal 3D scene representations, leading to suboptimal performance.
Therefore, we argue that incorporating 3D-aware \textit{visual pretext tasks} is crucial.
This allows models \textit{to be guided} to better understand both spatial relationships and comprehensive layouts.

To address this issue, we propose \textbf{\underline{r}}ec\textbf{\underline{o}}nstructive vi\textbf{\underline{s}}ual in\textbf{\underline{s}}truction tuning with \textbf{\underline{3D}}-awareness (\method), which introduces \textit{3D-aware vision-centric supervision signals} by adding a variety of vision-centric 3D-aware pretext tasks into the training procedure.
The high-level idea is presented in \Cref{fig:ours}, where we leverage input video frames to \textit{supervise those visual outputs} of LMMs directly.
To effectively inject 3D awareness, we consider two types of 3D-aware reconstructive visual pretext tasks, including cross-view reconstruction and global-view reconstruction.

(1) Cross-view reconstruction (the left part of \Cref{fig:ours}) is responsible for the detailed modeling of relationships between different views.
Specifically, it requires reconstruction on \textit{masked} views by analyzing overlapping information from other views. 
This process is crucial for tasks requiring fine-grained perception and precise alignment across various viewpoints such as 3D visual grounding~\cite{chen2020scanrefer, zhang2023multi3drefer}.

(2) Global-view reconstruction (the right part of \Cref{fig:ours}) contributes to the comprehensive understanding of the whole scene since it requires integrating information from all available perspectives to recover comprehensive Bird's-Eye View (BEV) images.
It synthesizes a complete and coherent overview of the entire scene, effectively capturing the full context and layout of the environment.
This approach is particularly useful for applications needing a comprehensive understanding such as 3D question-answering~\cite{ma2023sqa3d, azuma2022scanqa}.

With the aid of both (1) and (2), \method is equipped with a more accurate and effective 3D representation learning procedure.
Technically, a small denoising network~\cite{peebles2023dit} is responsible for reconstructing specific targets, \textit{i.e.}, masked views and BEV images for cross-view reconstruction and global-view reconstruction, respectively, conditioned on visual outputs from LMMs~\cite{wang2025ross}.


Empirically, as shown in \Cref{fig:radar}, \method outperforms previous approaches by a large margin across various evaluation protocols.
For instance, it achieves 63.0 EM on SQA3D~\cite{ma2023sqa3d}, 107.0 CIDEr on ScanQA~\cite{azuma2022scanqa}, 66.9 ROUGE on Scan2Cap~\cite{chen2021scan2cap}, 61.1 Acc@0.25 on ScanRefer~\cite{chen2020scanrefer}, and 59.6 F1@0.25 on Multi3DRefer~\cite{zhang2023multi3drefer}, outperforming the previous state-of-the-art Video-3D-LLM~\cite{zheng2024video3dllm} by +4.4 EM, +4.9 CIDEr, +5.2 ROUGE, +3.0 Acc@0.25, and +1.6 F1@0.25, respectively.
More importantly, towards the scarcity of high-quality 3D vision-language datasets, we leverage \method for semi-supervised learning by training on 50\% text-labeled data and applying the proposed 3D-aware visual objective to another 50\% unlabeled 3D vision-only data.
This approach even surpasses the 100\% text-supervised baseline in certain settings, demonstrating the significant potential of leveraging large amounts of unlabeled 3D data.
To summarize, our contributions are:
\begin{itemize}
    \item We introduce \method, a novel approach that enhances the ability of LMMs to understand 3D scenes on both spatial relationships and comprehensive layouts.
    \item We propose two distinct types of 3D-aware reconstructive visual pretext tasks: (1) cross-view reconstruction that focuses on modeling detailed relationships between different views, and (2) global-view reconstruction that provides a comprehensive understanding of the whole scene layout and context.
    \item \method brings significant improvements over previous state-of-the-art methods across multiple benchmarks, and effectively demonstrates the potential of leveraging large-scale unlabeled 3D visual data.
\end{itemize}
We hope this work will inspire future work in designing appropriate 3D-aware supervision signals for 3D LMMs.

\section{Related Works}\label{sec:related}

\noindent\textbf{3D Scene Understanding.}
As a fundamental requirement for embodied agents, 3D scene understanding has emerged as a focal point of research, witnessing numerous significant advancements over the years~\cite{durante2024agent, liang2023code, singh2023progprompt, brohan2023rt, black2024pi_0, zhu2024llava3d, zheng2024video3dllm, qi2025gpt4scene}, which have empowered embodied agents to accurately identify object positions, discern structures, and understand the relationships between objects within environments.
These advancements have been built upon foundational 3D perception tasks such as 3D visual grounding~\cite{achlioptas2020referit3d, chen2020scanrefer, chen2022language, huang2022multi}, 3D dense captioning~\cite{chen2021scan2cap, chen2023end, chen2024vote2cap}, and 3D question answering~\cite{azuma2022scanqa, ma2023sqa3d, yang2024thinking}, each demanding a comprehensive understanding of spatial positions and object relationships.
In contrast to conventional approaches~\cite{huang2022multi, chen2022language, guo2023viewrefer, jain2022bottom, luo20223d, wu2023eda, yang2021sat, yuan2022x, zhao20213dvg} that typically focus on specific tasks, our goal is to develop a generalist model, which is expected to address multiple aspects of 3D scene comprehension simultaneously interacting with humans using natural language.

\paragraph{Large Multimodal Models for Scene Understanding.}
The impressive generalization capabilities of state-of-the-art 2D large multimodal models (LMMs)~\cite{wang2025ross, li2024llava, wang2024qwen2vl, chen2024internvl2.5, wu2024deepseekvl2, zhang2024llavavideo, zhang2025videollama} has motivated researchers to adapt these models to understand 3D scenes~\cite{zhu2024llava3d, zheng2024video3dllm, qi2025gpt4scene, mei2024perla}.
A critical challenge in this adaptation is designing appropriate 3D scene representations that align well with the original 2D LMMs.
Prior attempts focus on utilizing features derived from 3D point clouds.
For instance, 3D-LLM~\cite{hong20233d} aggregates features from off-the-shelf 3D reconstruction backbones~\cite{jatavallabhula2019gradslam, jatavallabhula2023conceptfusion}.
PointLLM~\cite{xu2024pointllm} utilizes pre-trained 3D point cloud encoders, and LL3DA~\cite{chen2024ll3da} further leverages an extra Q-former~\cite{li2023blip} to extract useful information.
Several research~\cite{chen2024grounded, wang2023chat, huang2023leo, zhang2024chatscene} further enhance these methods by incorporating 3D detectors to provide object-centric representations.
However, these vision-only features are \textit{not} aligned with the feature space of LMMs.
To bridge this gap, LLaVA-3D~\cite{zhu2024llava3d} aggregates 2D-patch features from CLIP~\cite{radford2021learning} in the voxel space.
Video-3D-LLM~\cite{zheng2024video3dllm} treats multi-view images as video sequences and incorporates 3D information into video LMMs.
GPT4Scene~\cite{qi2025gpt4scene} improves it by introducing an extra BEV image.
This paper, on the basis of~\cite{zheng2024video3dllm}, regards multi-view images as videos and utilizes them as 3D scene representations, as it leverages the full potential of pre-trained 2D video LMMs and aligns more closely with human perception, where humans actually understand 3D scenes without explicit 3D point clouds.
Instead of input-level modifications, we aim to explore 3D-aware vision-centric designs for enhanced 3D spatial understanding capabilities.

\paragraph{Vision-Centric Designs in LMMs.}
Typical LMMs based on visual instruction tuning~\cite{liu2023visual, li2024llava, wang2024qwen2vl, wu2024deepseekvl2, chen2024internvl2.5} adopt a plug-in architecture, where pre-trained vision-language foundation models project images into visual tokens and subsequently serve as prefix tokens for multimodal comprehension.
This type of design is preliminary LLM-centric, as supervision solely comes from text tokens~\cite{wang2025ross}.
To address this limitation,~\cite{wang2025ross} pioneered the exploration of vision-centric supervision for 2D image LMMs.
This paper aims to extend this idea to 3D scene understanding.
However, this extension is non-trivial because it requires developing pretext tasks specific to 3D scenes rather than simply reconstructing original input images like~\cite{wang2025ross}.
Our exploration focuses on designing appropriate vision-centric pretext tasks to enhance 3D understanding within LMMs.

\paragraph{Visual Self-Supervised Learning.}
Visual self-supervised learning approaches rely on appropriate pretext tasks to extract scalable representations from large-scale data without any human annotations, along with representative studies in both images~\cite{wang2023hard, wang2023droppos, he2022masked, chen2020simple, he2020momentum}, videos~\cite{wang2023bootstrap, tong2022videomae, feichtenhofer2022masked, wei2022masked}, and 3D scenarios~\cite{achlioptas2018learning, yang2018foldingnet, gadelha2018multiresolution, jing2020self, feng2022disentangling}.
In the context of 3D perception, most previous methods have primarily concentrated on designing pretext tasks for point clouds~\cite{sauder2019self, huang2021spatio, sanghi2020info3d, xie2020pointcontrast, pang2022masked}.
In contrast, this work represents 3D scenes with multi-view images similar to~\cite{gao2023self, alwala2022pre}, which offers greater scalability compared to point clouds and is naturally aligned with existing 2D foundation models.

\begin{figure*}[t]
    \centering
    \includegraphics[width=1\linewidth]{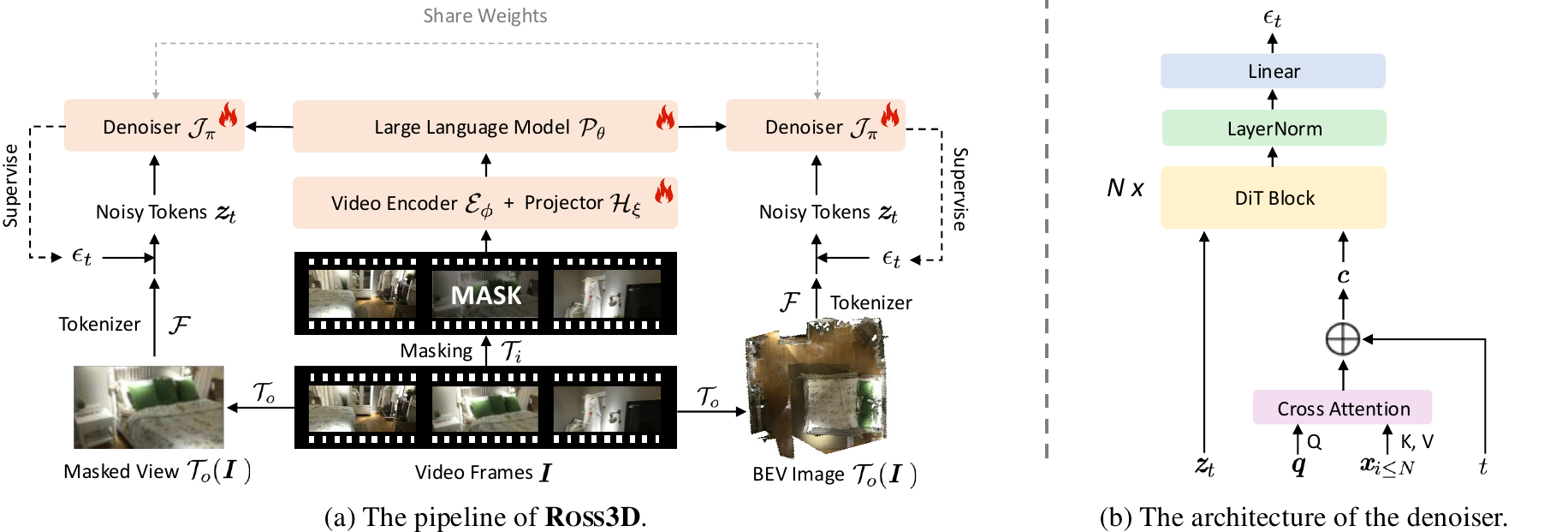}
    \vspace{-17pt}
    \caption{
    \textbf{Illustration of (a) \method and (b) the detailed architecture of the denoiser $\mathcal{J}_{\pi}$.}
    (a) Given raw video frames $\bm{I}$ for a 3D scene, we apply transformations to obtain inputs $\mathcal{T}_i(\bm{I})$ and targets $\mathcal{T}_o(\bm{I})$, respectively, and subsequently encourage LMMs to recover clean latent tokens $\bm{z}_0 = \mathcal{F} \circ \mathcal{T}_o(\bm{I})$ using noisy tokens $\bm{z}_t$ and visual outputs $\bm{x}_{i \leq N}$.
    (b) The denoiser is based on DiT~\cite{peebles2023dit}.
    Condition $\bm{c}$ is computed by a set of learnable queries $\bm{q}$, visual outputs $\bm{x}_{i \leq N}$, and timesteps $t$.
    }
    \label{fig:overview}
    \vspace{-10pt}
\end{figure*}

\section{Preliminaries}\label{sec:pre}

\textbf{Visual Instruction Tuning.}
Autoregressive LLMs model the canonical causal distribution of a text sentence $\bm{x} = \{\bm{x}_i\}_{i=1}^{T}$ as $p_{\theta} (\bm{x}) = p_{\theta}(\bm{x}_i | \bm{x}_{<i})$, where $\theta$ indicates trainable parameters and $T$ is the sequence length.
To comprehend visual signals $\bm{I}$, typical visual instruction tuning-based methods~\cite{liu2023visual} regard a sequence of visual features $\bm{v} = \mathcal{H}_{\xi} \circ \mathcal{E}_{\phi} (\bm{I})$ as prefix tokens, where $\mathcal{E}_{\phi}$ is a $\phi$-parameterized visual encoder such as CLIP~\cite{radford2021learning} and $\mathcal{H}_{\xi}$ is a $\xi$-parameterized multimodal projector.
Therefore, the canonical causal distribution for multimodal inputs becomes $p_{\Theta} = \prod_{i=1}^T p_{\Theta} (\bm{x}_i | \bm{x}_{<i}, v)$, where $\Theta = \{\theta, \phi, \xi\}$ indicates all parameters and $\bm{v} \in \mathbb{R}^{N \times D}$.
$N$ denotes the number of vision tokens and $D$ is the feature dimension.
The training objective is the standard cross-entropy, maximizing the log-likelihood of text outputs:
\begin{equation}\label{eq:lmm}
    \mathcal{L}_{\text{text}} (\bm{x}, \bm{I}; \Theta) = -\frac{1}{T-N} \sum_{i=N+1}^T \log p_{\Theta} (\bm{x}_i | \bm{x}_{<i}, \bm{v}),
\end{equation}
where \textit{only text outputs $\bm{x}_{i>N}$ are supervised}.

\paragraph{Reconstructive Visual Instruction Tuning.}
\cite{wang2025ross} introduced a simple yet effective reconstructive objective into the training procedure with enhanced fine-grained comprehension capabilities.
Specifically, it introduces an small $\pi$-parameterized denoising network $\mathcal{J}_{\pi}$ that is responsible to recover fine-grained tokens $\bm{z} = \mathcal{F} (\mathbf{I})$ conditioned on visual outputs $\bm{x}_{i\leq N}$, where $\mathcal{F}$ is the teacher tokenizer such as VAE~\cite{kingma2013vae}.
This type of 2D objective is 
\begin{equation}\label{eq:ross}
    \mathcal{L}_{\text{2D}} (\bm{x}, \bm{I}; \Theta) = \mathbb{E}_{t, \bm{\epsilon}} \left[ || \mathcal{J}_{\pi} (\bm{z}_t | \bm{x}_{i\leq N}, t) - \bm{\epsilon}||^2 \right].
\end{equation}
While this objective is empirically effective in 2D understanding, it does not introduce any 3D awareness.

\section{Method}\label{sec:method}

We propose a 3D generalist model for indoor scene understanding, namely \method.
In contrast to previous approaches that focus on injecting 3D information into 2D LMMs through input-level modifications, \method is equipped with novel 3D-aware vision-centric supervision signals.
In the following, we elaborate on our \method step by step.
First, we provide a comprehensive overview of our method in \Cref{sec:overview}.
Subsequently, implementation details of two proposed 3D-aware pretext tasks are provided in \Cref{sec:cross} and \Cref{sec:global}.
Finally, detailed formulations of training objectives are introduced in \Cref{sec:objective}.

\subsection{Overview}\label{sec:overview}

An overview of our \method is presented in \Cref{fig:overview}\textcolor{red}{a}, which contains a video encoder $\mathcal{E}_{\phi}$, a large language model $\mathcal{P}_{\theta}$, and a denoiser $\mathcal{J}_{\pi}$.
Different from conventional methods~\cite{zheng2024video3dllm, hong20233d, zhu2024llava3d, mei2024perla, qi2025gpt4scene, zhang2024chatscene, huang2023leo} that solely supervise text outputs $\bm{x}_{i > N}$, we design a series of 3D-aware vision-centric supervision $\mathcal{L}_{\text{3D}}$ for visual outputs $\bm{x}_{i \leq N}$.
\begin{equation}\label{eq:ross3d}
    \mathcal{L}_{\text{3D}} (\bm{x}, \bm{I}; \Theta) = \mathcal{D} (\mathcal{J}_{\pi} (\bm{x}_{i \leq N}), \mathcal{F} \circ \mathcal{T}_o(\bm{I})),
\end{equation}
where $\mathcal{D}$ is a specific distance metric, and this paper takes the diffusion denoising process by default.

Transformations at the input-level  $\mathcal{T}_i$ and the output-level $\mathcal{T}_o$ are applied to video frames $\bm{I}$ to obtain input videos and reconstruction targets, respectively.
LMMs are required to recover $\mathcal{T}_o(\bm{I})$ while taking $\mathcal{T}_i(\bm{I})$ as inputs.
In particular, \Cref{eq:ross3d} degenerates into ``vanilla reconstruction'', \textit{i.e.}, \Cref{eq:ross}, \textit{without} 3D-awareness when both $\mathcal{T}_i$ and $\mathcal{T}_o$ are identity functions, \textit{i.e.}, reconstruction targets are identical to inputs.
Therefore, to inject 3D awareness, design choices of transformations are crucial.

In the following, we introduce how we choose appropriate transformations to conduct \textit{3D-aware} self-supervised pretext tasks, including cross-view reconstruction in \Cref{sec:cross} and global-view reconstruction in \Cref{sec:global}.

Please note that the actual inputs for \method include video frames alongside a corresponding depth map for each frame to produce position-aware video representations discussed in the \supp. 
We omit the depth inputs in this section as this simplification helps to focus on the essential aspects and does not influence our motivation or the overall pipeline.

\subsection{Cross-View Reconstruction}\label{sec:cross}

Cross-view reconstruction enables reconstructing masked views based on other views, contributing to enhanced modeling of relationships between different views, which is crucial for tasks requiring fine-grained perception and precise alignment across various viewpoints such as 3D visual grounding~\cite{zhang2023multi3drefer, chen2020scanrefer}.

In general, input transformation $\mathcal{T}_i$ indicates randomly masking a subset of views, and output transformation $\mathcal{T}_o$ is obtaining those masked views.
Formulaly, given multi-view images $\bm{I} \in \mathbb{R}^{M \times H \times W \times 3}$, where $M$ indicates the number of views and $(H, W)$ is the spatial resolution, we first generate a \textit{view-aware} binary mask $\bm{M} \in \{0, 1\}^{M}$ with a mask ratio $\gamma$, \textit{i.e.}, $\sum_{i=j}^M \bm{M}_j = (1 - \gamma) M$, where 1 means unmasked views while 0 indicates the opposite.
Features of those masked views are subsequently replaced with learnable mask tokens $\bm{m} \in \mathbb{R}^{D}$.
Specifically, the encoded visual feature $\bm{v}_i$ for each frame $i$ becomes:
\begin{equation}\label{eq:cross_v}
    \bm{v}_j = \left\{
    \begin{aligned}
        &\mathcal{H}_{\xi} \circ \mathcal{E}_{\phi} (\bm{I}_j), &\text{if}\ \ \bm{M}_j = 1, \\
        &\bm{m}_j, &\text{otherwise.}
    \end{aligned}
    \right.
\end{equation}
Similarly, reconstruction targets $\bm{z}_0$ becomes the latent tokens provided by the teacher $\mathcal{F}$ for those \textit{masked} views:
\begin{equation}\label{eq:cross_z}
    \bm{z}_0 = \{\mathcal{F}(\bm{I}_j) \mid \bm{M}_j=0\}.
\end{equation}

Therefore, the formulation of cross-view reconstruction can be obtained through rewriting \Cref{eq:ross3d} by
\begin{equation}
\begin{aligned}
    \mathcal{L}_{\text{3D}}^{\text{cross}} 
    &= \frac{1}{\gamma M}\sum_{j=1}^M (1-\bm{M}_j) \cdot \mathcal{D}(\mathcal{J}_{\pi} \circ \mathcal{P}_{\theta}(\bm{v}), \mathcal{F}(\bm{I}_j)),
\end{aligned}
\end{equation}
where $\bm{v}$ indicates visual features defined in \Cref{eq:cross_v}.
Here, language tokens are omitted as they are always behind visual tokens. Thus, they do not influence the forward motion of visual parts due to their causal nature.

\paragraph{Discussion.}
As masking may lead to discrepancies between training and testing, we apply this objective every $\Delta t$ steps and we set $\Delta t=4$ by default.
For the same reason, a relatively small mask ratio, \textit{e.g.}, 25\%, is also important.

\subsection{Global-View Reconstruction}\label{sec:global}

Global-view reconstruction enables reconstructing the BEV image of the whole scene, resulting in an improved understanding of the environment, which is crucial for tasks requiring comprehensive comprehension such as 3D question-answering~\cite{ma2023sqa3d, azuma2022scanqa}.

Under this case, $\mathcal{T}_i$ is the same as cross-view reconstruction, while $\mathcal{T}_o$ is responsible for converting inputs into a BEV image $\bm{I}_{\text{BEV}}$ using both egocentric video, extrinsic parameters for each frame, and the corresponding camera intrinsic matrix.
We use 3D reconstruction techniques to generate 3D meshes and point clouds, and render a BEV image from the top-down view.

Formally, the objective of global-view reconstruction can be obtained through rewriting \Cref{eq:ross3d} by
\begin{equation}
\begin{aligned}
    \mathcal{L}_{\text{3D}}^{\text{global}} (\bm{x}, \bm{I}; \Theta) 
    &= \mathcal{D}(J_{\pi} \circ \mathcal{P}_{\theta} (\bm{v}), \mathcal{F}(\bm{I}_{\text{BEV}})).
\end{aligned}
\end{equation}

\noindent\textbf{Discussion.}
Since BEV images are actually rendered from sparse scene point clouds, this process can result in numerous black blocks. 
Therefore, we simply filter out these blank spaces during reconstruction.

\subsection{Training Objectives}\label{sec:objective}

Following~\cite{wang2025ross}, we leverage a simple denoising objective, as vanilla regression may suffer from heavy spatial redundancy of visual signals, and thus fail to produce meaningful supervision for LMMs.
Technically, as demonstrated in \Cref{fig:overview}\textcolor{red}{a}, our \method regards high-level visual outputs $\bm{x}_{i \leq N}$ as conditions to recover clean latent tokens $\bm{z}_0$ from noisy tokens $\bm{z}_t$.
By default, we take a continuous VAE~\cite{kingma2013vae} regularized by Kullback–Leibler (KL) divergence provided by FLUX~\cite{flux2024} since it is believed to capture sufficient image details.
The training follows a diffusion process~\cite{ho2020denoising}:
\begin{equation}
    \mathcal{D}(\mathcal{J}_{\pi} \circ \mathcal{P}_{\theta} (\bm{v}), \bm{z}_0) = \mathbb{E}_{t, \bm{\epsilon}} \left[ || \mathcal{J}_{\pi} (\bm{z}_t | \mathcal{P}_{\theta} (\bm{v}), t) - \bm{\epsilon}||^2 \right],
\end{equation}
where $\bm{z}_t$ is sampled from $\mathcal{N}(\sqrt{1-\beta_t} \bm{z}_{t-1}, \beta_t \bm{1})$ and $\bm{1}$ here indicates the identity matrix.
Following~\cite{ho2020denoising}, $\bm{z}_t$ could be sampled directly from $\bm{z}_0$ by letting $\bar{\alpha}_t = \prod_{i=1}^t (1-\beta_i)$
\begin{equation}
    \bm{z}_t = \sqrt{\bar{\alpha}_t} \bm{z}_0 + \sqrt{1 - \bar{\alpha}_t} \bm{\epsilon}, \quad \bm{\epsilon} \sim \mathcal{N}(\bm{0}, \bm{1}).
\end{equation}


Other objectives include standard cross-entropy loss introduced in \Cref{eq:lmm} and grounding loss described later in the \supp.

\begin{table*}[t]
    \centering\small
    \setlength{\tabcolsep}{8.57pt}
    \begin{tabular}{lcc cc ccccc}
    \toprule
    \multirow{2}{*}{Method} & \multirow{2}{*}{\begin{tabular}[c]{@{}c@{}}Point\\ Encoder\end{tabular}} & \multirow{2}{*}{\begin{tabular}[c]{@{}c@{}}Vision\\ Encoder\end{tabular}} & \multicolumn{2}{c}{SQA3D$_{\text{test}}$} & \multicolumn{5}{c}{ScanQA$_{\text{val}}$} \\
    \cmidrule(lr){4-5}
    \cmidrule(lr){6-10}
    & & & EM & EM-R & CIDEr & BLEU-4 & METEOR & ROUGE & EM \\
    \midrule
    \multicolumn{10}{l}{\textit{Expert Models}} \\
    SQA3D~\cite{ma2023sqa3d} & \checkmark & -- & 46.6 & -- & -- & -- & -- & -- & -- \\
    ScanQA~\cite{azuma2022scanqa} & \checkmark & -- & -- & -- & 64.9 & 10.1 & 13.1 & 33.3 & 21.1 \\
    3D-VLP~\cite{jin2023context} & \checkmark & -- & -- & -- & -- & 11.2 & 13.5 & 34.5 & 21.7 \\
    3D-VisTA~\cite{zhu20233dvista} & \checkmark & -- & -- & -- & -- & -- & 13.9 & 35.7 & 22.4 \\
    \midrule
    \multicolumn{10}{l}{\textit{2D LMMs}} \\
    InternVL2-8B~\cite{internvl2} & -- & \checkmark & 33.0 & 45.3 & 62.5 & 3.3 & 14.5 & 34.3 & -- \\
    Qwen2-VL-7B~\cite{wang2024qwen2vl} & -- & \checkmark & 40.7 & 46.7 & 53.9 & 3.0 & 11.4 & 29.3 & -- \\
    LLaVA-Video-7B~\cite{zhang2024llavavideo} & -- & \checkmark & 48.5 & -- & 88.7 & 3.1 & 17.7 & 44.6 & -- \\
    \midrule
    \multicolumn{10}{l}{\textit{3D LMMs}} \\
    Chat-3D~\cite{wang2023chat} & \checkmark & -- & -- & -- & 53.2 & 6.4 & 11.9 & 28.5 & -- \\
    3D-LLM~\cite{hong20233d} & \checkmark & \checkmark & -- & -- & 69.4 & 12.0 & 14.5 & 35.7 & 20.5 \\
    Scene-LLM~\cite{fu2024scene} & \checkmark & \checkmark & 53.6 & -- & 80.0 & 11.7 & 15.8 & 35.9 & 27.2 \\
    LL3DA~\cite{chen2024ll3da} & \checkmark & -- & -- & -- & 76.8 & -- & 15.9 & 37.3 & -- \\
    LEO~\cite{huang2023leo} & \checkmark & \checkmark & 50.0 & 52.4 & 80.0 & 11.5 & 16.2 & 39.3 & 21.5 \\
    ChatScene~\cite{zhang2024chatscene} & \checkmark & \checkmark & 54.6 & 57.5 & 87.7 & 14.3 & 18.0 & 41.6 & 21.6 \\
    Grounded 3D-LLM~\cite{chen2024grounded} & \checkmark & \checkmark & -- & -- & 72.7 & 13.4 & -- & -- & -- \\
    LLaVA-3D~\cite{zhu2024llava3d} & -- & \checkmark & 55.6 & 57.6 & 91.7 & 14.5 & 20.7 & 50.1 & 27.0 \\
    Video-3D-LLM~\cite{zheng2024video3dllm} & -- & \checkmark & 58.6 & -- & 102.1 & 16.4 & 20.0 & 49.3 & 30.1 \\
    \textcolor{codegray}{GPT4Scene-HDM$^{\ddag}$}~\cite{qi2025gpt4scene} & \textcolor{codegray}{--} & \textcolor{codegray}{\checkmark} & \textcolor{codegray}{59.4} & \textcolor{codegray}{62.4} & \textcolor{codegray}{96.3} & \textcolor{codegray}{15.5} & \textcolor{codegray}{18.9} & \textcolor{codegray}{46.5} & \textcolor{codegray}{--} \\
    \rowcolor{Light}
    \method & -- & \checkmark & \textbf{63.0} & \textbf{65.7} & \textbf{107.0} & \textbf{17.9} & \textbf{20.9} & \textbf{50.7} & \textbf{30.8} \\
    \bottomrule
    \end{tabular}
    \vspace{-5pt}
    \caption{
    \textbf{Evaluation of 3D question-answering} on SQA3D~\cite{ma2023sqa3d} and ScanQA~\cite{azuma2022scanqa}.
    ``Expert models'' are customized for specific tasks with task-oriented decoders. 
    General 2D LMMs~\cite{internvl2, wang2024qwen2vl, zhang2024llavavideo} are evaluated in a zero-shot setting.
    ``EM'' stands for top-1 exact match and ``EM-R'' means the refined exact match following~\cite{huang2023leo}.
    ``--'' indicates the number is not available for us.
    ``$\ddag$'' indicates this result is achieved by adopting a larger input resolution, \textit{i.e.}, 512$\times$490, and incorporating extra BEV inputs.
    }
    \label{tab:3dqa}
    \vspace{-10pt}
\end{table*}

\section{Experiments}\label{sec:exp}


\noindent\textbf{Datasets.}
To evaluate the 3D scene understanding capabilities of our \method, we conduct experiments across five representative benchmarks, including SQA3D~\cite{ma2023sqa3d} for situated reasoning, ScanQA~\cite{azuma2022scanqa} for spatial understanding, Scan2Cap~\cite{chen2021scan2cap} for captioning specific objects, ScanRefer~\cite{chen2020scanrefer} and Multi3DRefer~\cite{zhang2023multi3drefer} for detecting objects in single-target and multiple-target scenarios.
All these datasets are derived from ScanNet~\cite{dai2017scannet}, which is an extensively annotated collection of RGB-D video data, encompassing 1,513 scans of 3D indoor scenes.
%
%

\paragraph{Metrics.}
Widely used evaluation metrics are utilized for
each benchmark.
For SQA3D~\cite{ma2023sqa3d}, we evaluate
the performance using exact match accuracy (EM) and the refined exact match protocol (EM-R) following~\cite{huang2023leo, zhu2024llava3d}.
For ScanQA~\cite{azuma2022scanqa}, we use EM, BLEU-4~\cite{papineni2002bleu}, METEOR~\cite{banerjee2005meteor}, ROUGE~\cite{lin2004rouge}, and CIDEr~\cite{vedantam2015cider}.
For Scan2Cap~\cite{chen2021scan2cap}, we combine captioning metrics (CIDEr, BLEU-4, METEOR, and ROUGE) with an IoU threshold of 0.5 between predicted and reference bounding boxes.
For ScanRefer~\cite{chen2020scanrefer}, we report Acc@0.25 and Acc@0.5, where a prediction is considered correct only if the IoU exceeds 0.25 and 0.5, respectively.
For Multi3DRefer~\cite{zhang2023multi3drefer}, we combine F1 scores and IoU thresholds, \textit{i.e.}, F1@0.25 and F1@0.5.

\begin{table}[t]
    \centering\small
    \setlength{\tabcolsep}{2.6pt}
    \begin{tabular}{l cccc}
    \toprule
    \multirow{2}{*}{Method} & \multicolumn{4}{c}{Scan2Cap$_{\text{val}}$ (IoU@0.5)} \\
    \cmidrule(lr){2-5}
    & ROUGE & BLEU-4 & METEOR & CIDEr \\
    \midrule
    \multicolumn{5}{l}{\textit{Expert Models}} \\
    Scan2Cap~\cite{chen2021scan2cap} & 44.5 & 23.3 & 22.0 & 35.2 \\
    3DJCG~\cite{cai20223djcg} & 50.8 & 31.0 & 24.2 & 49.5 \\
    3D-VLP~\cite{jin2023context} & 51.5 & 32.3 & 24.8 & 54.9 \\
    3D-VisTA~\cite{zhu20233dvista} & 54.3 & 34.0 & 26.8 & 61.6 \\
    \midrule
    \multicolumn{5}{l}{\textit{3D LMMs}} \\
    LL3DA~\cite{chen2024ll3da} & 55.1 & 36.8 & 26.0 & 65.2 \\
    LEO~\cite{huang2023leo} & 58.1 & 38.2 & 27.9 & 72.4 \\
    ChatScene~\cite{zhang2024chatscene} & 58.1 & 36.3 & -- & 77.1 \\
    LLaVA-3D~\cite{zhu2024llava3d} & 63.4 & 41.1 & 30.2 & 79.2 \\
    Video-3D-LLM~\cite{zheng2024video3dllm} & 62.3 & 42.4 & 28.9 & \textbf{83.8} \\
    \textcolor{codegray}{GPT4Scene-HDM$^{\ddag}$}~\cite{qi2025gpt4scene} & \textcolor{codegray}{59.3} & \textcolor{codegray}{40.6} & \textcolor{codegray}{--} & \textcolor{codegray}{--} \\
    \rowcolor{Light}
    \method & \textbf{66.9} & \textbf{43.4} & \textbf{30.3} & 81.3 \\
    \bottomrule
    \end{tabular}
    \vspace{-5pt}
    \caption{
    \textbf{Evaluation of 3D dense captioning} on Scan2Cap \cite{chen2021scan2cap}.
    ``$\ddag$'' indicates this result is achieved by adopting a larger input resolution, \textit{i.e.}, 512$\times$490, and incorporating extra BEV inputs.
    %
    }
    \label{tab:scan2cap}
    \vspace{-10pt}
\end{table}

\paragraph{Implementation Details.}
We build our \method based on LLaVA-Video-7B~\cite{zhang2024llavavideo}, which is then fine-tuned on the combination of training sets of SQA3D~\cite{ma2023sqa3d}, ScanQA~\cite{azuma2022scanqa}, Scan2Cap~\cite{chen2021scan2cap}, ScanRefer~\cite{chen2020scanrefer}, and Multi3DRefer~\cite{zhang2023multi3drefer} for one epoch, using the AdamW optimizer with a global batch size of 256.
The learning rates peak at 1e-5 for the LLM during the warmup phase and the vision encoder is kept frozen.
All experiments are conducted with 8$\times$A100-80G.
Each scene is represented by 32 frames, with the resolution of each frame being 384$\times$384.
BEV images are rendered from point clouds with a resolution of 432$\times$432.

\subsection{Comparison with State-of-the-Arts}\label{sec:sota}

\noindent\textbf{Comparison Alternatives.}
We include both expert models designed for specific tasks and LMM-based models with different input representations.
\textit{Expert models} include ScanQA~\cite{azuma2022scanqa}, Scan2Cap~\cite{chen2021scan2cap}, ScanRefer~\cite{chen2020scanrefer}, MVT~\cite{huang2022multi}, 3DVG-Trans~\cite{zhao20213dvg}, ViL3DRel~\cite{chen2022language}, M3DRef-CLIP~\cite{zhang2023multi3drefer}, 3D-VLP~\cite{jin2023context}, 3DJCG~\cite{cai20223djcg}, and 3D-VisTA~\cite{zhu20233dvista}.
\textit{2D LMMs} include InternVL2~\cite{internvl2}, Qwen2-VL~\cite{wang2024qwen2vl}, and LLaVA-Video~\cite{zhang2024llavavideo}.
\textit{3D LMMs} include 3D-LLM~\cite{hong20233d} that leverages 2D encoders pre-trained 3D tasks, Scene-LLM~\cite{fu2024scene} and LL3DA~\cite{chen2024ll3da} that utilize point cloud features, Chat-3D~\cite{wang2023chat}, LEO~\cite{huang2023leo}, ChatScene~\cite{zhang2024chatscene}, and Grounded 3D-LLM~\cite{chen2024grounded} that incoperate object representations, LLaVA-3D~\cite{zhu2024llava3d} that aggregates 2D features in the 3D voxel space, and Video-3D-LLM~\cite{zheng2024video3dllm} and GPT4Scene~\cite{qi2025gpt4scene} that treat multi-view images as video sequences.

\paragraph{3D Question Answering.}
We compare our \method with other methods on 3D question answering in \Cref{tab:3dqa}.
As demonstrated in the table, \method achieves 63.0 EM on SQA3D~\cite{ma2023sqa3d} and 107.0 CIDEr on ScanQA~\cite{azuma2022scanqa}, outperforming previous state-of-the-art Video-3D-LLM~\cite{zheng2024video3dllm} by +4.4 EM on SQA3D~\cite{ma2023sqa3d} and +4.9 CIDEr on ScanQA~\cite{azuma2022scanqa}.

\paragraph{3D Dense Captioning.}
We compare our \method with other methods on 3D dense captioning in \Cref{tab:scan2cap}.
As demonstrated in the table, \method achieves 66.9 ROUGE, 43.4 BLEU-4, and 30.3 METEOR  on Scan2Cap~\cite{chen2021scan2cap}, outperforming previous state-of-the-art Video-3D-LLM~\cite{zheng2024video3dllm} by +4.6 ROUGE, +1.0 BLEU-4, and +1.4 METEOR.

\paragraph{3D Visual Grounding.}
We compare \method with other methods on 3D visual grounding in \Cref{tab:grounding}.
As demonstrated in the table, \method achieves 61.1 Acc@0.25 and 54.4 Acc@0.5 on ScanRefer~\cite{chen2020scanrefer}, and 59.6 F1@0.25 and 54.3 F1@0.5 on Multi3DRefer~\cite{zhang2023multi3drefer}, outperforming previous state-of-the-art Video-3D-LLM~\cite{zheng2024video3dllm} by +3.0 Acc@0.25 and +2.7 Acc@0.5 on ScanRefer~\cite{chen2020scanrefer}, and +1.6 F1@0.25 and +1.6 F1@0.5 on Multi3DRefer~\cite{zhang2023multi3drefer}, respectively.

\subsection{Ablation Studies}\label{sec:ablation}

In this section, we report EM for SQA3D~\cite{ma2023sqa3d} , CIDEr for ScanQA~\cite{azuma2022scanqa}, Acc@0.25 for ScanRefer~\cite{chen2020scanrefer}, and F1@0.25 for Multi3DRefer~\cite{zhang2023multi3drefer} by default.

\paragraph{Effectiveness of Each 3D-Aware Pretext Task.}
We study the effectiveness of each 3D-aware pretext task using different input representations in \Cref{tab:abl_each_task}.
Specifically, we compare our proposed two \textit{3D-aware} tasks, \textit{i.e.}, cross-view reconstruction and global-view reconstruction, with vanilla reconstruction \textit{without} 3D-awareness, and visual instruction tuning baselines.
Input representations include scene-level 3D features provided by 3D-LLM~\cite{hong20233d} and position-aware video representations proposed by Video-3D-LLM~\cite{zheng2024video3dllm}.
According to~\cite{hong20233d}, these 3D point cloud features are obtained by (1) extracting object masks using Mask2Former~\cite{cheng2022mask2former} and SAM~\cite{kirillov2023segment}, (2) extracting features of each object using BLIP-2~\cite{li2023blip}, and (3) reconstructing 3D features from extracted multi-view 2D features.
Following the official implementation of 3D-LLM~\cite{hong20233d}, we load the v2 pre-trained model and fine-tune on each task \textit{separately} for 100 epochs.
We fail to conduct experiments on ScanRefer~\cite{chen2020scanrefer} with 3D-LLM~\cite{hong20233d} as this part of fine-tuning code is unavailable.

\begin{table}[t]
    \centering\small
    \setlength{\tabcolsep}{1.3pt}
    \begin{tabular}{l cccc}
    \toprule
    \multirow{2}{*}{Method} & \multicolumn{2}{c}{ScanRefer$_{\text{val}}$} & \multicolumn{2}{c}{Multi3DRefer$_{\text{val}}$} \\
    \cmidrule(lr){2-3}
    \cmidrule(lr){4-5}
    & Acc@0.25 & Acc@0.5 & F1@0.25 & F1@0.5 \\
    \midrule
    \multicolumn{5}{l}{\textit{Expert Models}} \\
    ScanRefer~\cite{chen2020scanrefer} & 37.3 & 24.3 & -- & -- \\
    MVT~\cite{huang2022multi} & 40.8 & 33.3 & -- & -- \\
    3DVG-Trans~\cite{zhao20213dvg} & 47.6 & 34.7 & -- & 25.5 \\
    ViL3DRel~\cite{chen2022language} & 47.9 & 37.7 & -- & -- \\
    3DJCG~\cite{cai20223djcg} & 49.6 & 37.3 & -- & 26.6 \\
    M3DRef-CLIP~\cite{zhang2023multi3drefer} & 51.9 & 44.7 & 42.8 & 38.4 \\
    \midrule
    \multicolumn{5}{l}{\textit{3D LMMs}} \\
    3D-LLM~\cite{hong20233d} & 30.3 & -- & -- & -- \\
    Ground 3D-LLM~\cite{chen2024grounded} & 47.9 & 44.1 & 45.2 & 40.6 \\
    ChatScene~\cite{zhang2024chatscene} & 55.5 & 50.2 & 57.1 & 52.4 \\
    LLaVA-3D~\cite{zhu2024llava3d} & 54.1 & 42.4 & -- & -- \\
    Video-3D-LLM~\cite{zheng2024video3dllm} & 58.1 & 51.7 & 58.0 & 52.7 \\
    \textcolor{codegray}{GPT4Scene-HDM$^{\ddag}$}~\cite{qi2025gpt4scene} & \textcolor{codegray}{62.6} & \textcolor{codegray}{57.0} & \textcolor{codegray}{64.5} & \textcolor{codegray}{59.8} \\
    \rowcolor{Light}
    \method & \textbf{61.1} & \textbf{54.4} & \textbf{59.6} & \textbf{54.3} \\
    \bottomrule
    \end{tabular}
    \vspace{-5pt}
    \caption{
    \textbf{Evaluation of 3D visual grounding} on ScanRefer \cite{chen2020scanrefer} and Multi3DRefer~\cite{zhang2023multi3drefer}.
    ``$\ddag$'' indicates this result is achieved by adopting a larger input resolution, \textit{i.e.}, 512$\times$490, $\approx$1.7$\times$ pixels than ours, and incorporating extra BEV inputs.
    }
    \label{tab:grounding}
\end{table}

\begin{table}[t]
    \centering\small
    \setlength{\tabcolsep}{4pt}
    \begin{tabular}{rl lll}
    \toprule
    & Method & SQA3D & ScanQA & ScanRefer \\
    \midrule
    \multicolumn{5}{l}{\textit{Point-based Representation}} \\
    {\scriptsize{1}} & 3D-LLM$^{\ddag}$~\cite{hong20233d} & 49.4 & 68.9 & -- \\
    {\scriptsize{2}} & \textcircled{\scriptsize{1}} + vanilla & 48.9 \down{0.5} & 68.9 \textcolor{codegray}{-- 0.0} & -- \\
    {\scriptsize{3}} & \textcircled{\scriptsize{1}} + cross-view   & 51.0 \up{1.6} & 70.3 \up{0.4} & -- \\
    {\scriptsize{4}} & \textcircled{\scriptsize{1}} + global-view   & \underline{54.3 \up{4.9}} & \underline{71.0 \up{1.1}} & -- \\
    {\scriptsize{5}} & \textcircled{\scriptsize{1}} + \textcircled{\scriptsize{3}} + \textcircled{\scriptsize{4}} & \textbf{55.0 \up{5.6}} & \textbf{73.1 \up{2.2}} & -- \\
    \midrule
    \multicolumn{5}{l}{\textit{Video-based Representation}} \\
    {\scriptsize{6}} & Video-3D-LLM~\cite{zheng2024video3dllm} & 58.6 & 102.1 & 58.1 \\
    {\scriptsize{7}} & \textcircled{\scriptsize{6}} + vanilla & 58.8 \textcolor{codegray}{-- 0.0} & 103.5 \up{1.4} & 58.2 \up{0.1} \\
    {\scriptsize{8}} & \textcircled{\scriptsize{6}} + cross-view & 60.0 \up{1.4} & 103.6 \up{1.5} & 60.3 \up{2.1} \\
    {\scriptsize{9}} & \textcircled{\scriptsize{6}} + global-view & 61.6 \up{3.0} & 105.6 \up{3.5} & 58.8 \up{0.7}  \\
    \rowcolor{Light}
    {\scriptsize{10}} & \textcircled{\scriptsize{6}} + \textcircled{\scriptsize{8}} + \textcircled{\scriptsize{9}} & \textbf{63.0 \up{4.4}} & \textbf{107.0 \up{4.9}} & \textbf{61.1 \up{3.0}} \\
    \bottomrule
    \end{tabular}
    \vspace{-5pt}
    \caption{
    \textbf{Ablations on 3D-aware pretext tasks} with different input representations, including point-based~\cite{hong20233d}, and video-based~\cite{zheng2024video3dllm}.
    ``Vanilla'' indicates directly reconstruction \textit{without} 3D-awareness.
    Our default setting is \colorbox{Light}{highlighted} in color.
    ``$\ddag$'' means our reproduction using the official code.
    ``--'' indicates this part of code is unavailable for us.
    }
    \label{tab:abl_each_task}
    \vspace{-10pt}
\end{table}

As demonstrated in the table, we can draw the following three important conclusions.
(1) Pretext tasks \textit{with} 3D-awareness is crucial, as vanilla reconstruction brings marginal improvements.
(2) Each 3D-aware pretext task is effective.
(3) The proposed two pretext tasks promote each other, contributing to significant improvements \textit{across different input representations}.
Moreover, cross-view reconstruction is particularly effective for 3D visual grounding on ScanRefer~\cite{chen2020scanrefer} while global-view reconstruction is effective for 3D question answering on~\cite{ma2023sqa3d, azuma2022scanqa}.

\paragraph{\method \textit{v.s.} Adding 3D Features.}
In \Cref{tab:abl_encoder}, we compare our output-level supervision solution with the input-level aggregation alternative.
Specifically, the second row in \Cref{tab:abl_encoder} indicates we aggregate scene-level 3D point cloud features provided by~\cite{hong20233d} with original position-aware video representations.
This table demonstrates that our \method is much more effective than simply adding 3D features.

\begin{table}[t]
    \centering\small
    \setlength{\tabcolsep}{4.4pt}
    \begin{tabular}{rl lll}
    \toprule
    & Method & SQA3D & ScanQA & ScanRefer \\
    \midrule
    {\scriptsize{1}} & Video-3D-LLM~\cite{zheng2024video3dllm} & 58.6 & 102.1 & 58.1 \\
    {\scriptsize{2}} & \textcircled{\scriptsize{1}} + 3D features & 59.1 \up{0.5} & 102.4 \up{0.3} & 57.8 \down{0.3} \\
    \rowcolor{Light}
    {\scriptsize{3}} & \method & \textbf{63.0 \up{4.4}} & \textbf{107.0 \up{4.9}} & \textbf{61.1 \up{3.0}} \\
    \bottomrule
    \end{tabular}
    \vspace{-5pt}
    \caption{
    \textbf{\method \textit{v.s.} Adding 3D Features.}
    Scene-level 3D features are provided by~\cite{hong20233d}, which are aggregated with video features via cross attention.
    }
    \label{tab:abl_encoder}
\end{table}

\paragraph{Semi-Supervised Learning with \method.}
As high-quality 3D vision-language data is quite limited, scaling up 3D LMMs poses a significant challenge. 
To address this issue by \textit{leveraging knowledge directly from raw 3D visual data}, we conduct the following \textit{semi-supervised} experiments in \Cref{tab:semi}, as \method naturally allows learning from 3D sequences \textit{without} text annotations.
Specifically, we split the training set into two non-overlapping parts and then run 4 settings: \textcircled{\scriptsize{1}} training with 50\% text data with conventional $\mathcal{L}_{\text{text}}$, \textcircled{\scriptsize{2}} training with 50\% text data with visual supervision provided by \method, \textit{i.e.}, with $\mathcal{L}_{\text{3D}}$ in the table, \textcircled{\scriptsize{3}} training 50\% with text data and applying \method on the other 50\% \textit{without} text annotations, and \textcircled{\scriptsize{4}} training with 100\% text data.
As demonstrated in the table, \textcircled{\scriptsize{3}} significantly outperforms both \textcircled{\scriptsize{1}} and \textcircled{\scriptsize{2}}.
It even surpasses the supervised upper bound represented by \textcircled{\scriptsize{4}} on ScanQA (103.2 \textit{v.s.} 102.1 on CIDEr).
This result highlights the effectiveness of $\mathcal{L}_{\text{3D}}$ in learning \textit{directly} from visual signals.

\begin{table}[t]
    \centering\small
    \setlength{\tabcolsep}{3pt}
    \begin{tabular}{r lllll}
    \toprule
    & 50\% Data &
    + 50\% Data & SQA3D & ScanQA & ScanRefer \\
    \midrule
    \scriptsize{1} & $\mathcal{L}_{\text{text}}$ & -- & 55.1 & 100.3 & 57.0 \\
    \scriptsize{2} & $\mathcal{L}_{\text{text}}$ + $\mathcal{L}_{\text{3D}}$ & -- & 56.4 \up{1.3} & 101.7 \up{0.6} & 57.4 \up{0.4} \\
    \rowcolor{Light}
    \scriptsize{3} & $\mathcal{L}_{\text{text}}$ + $\mathcal{L}_{\text{3D}}$ & $\mathcal{L}_{\text{3D}}$ & \textbf{57.7 \up{2.6}} & \textbf{103.2 \up{2.9}} & \textbf{57.9 \up{0.9}} \\
    \textcolor{codegray}{\scriptsize{4}} & \textcolor{codegray}{$\mathcal{L}_{\text{text}}$} & \textcolor{codegray}{$\mathcal{L}_{\text{text}}$} & \textcolor{codegray}{58.6} & \textcolor{codegray}{102.1} & \textcolor{codegray}{58.1} \\
    \bottomrule
    \end{tabular}
    \vspace{-5pt}
    \caption{
    \textbf{Semi-supervised learning with \method.}
    ``\checkmark'' indicates applying the particular objective on this part of data, while ``--'' means the opposite.
    \textcircled{\scriptsize{4}} is actually the standard Video-3D-LLM~\cite{zheng2024video3dllm} baseline.
    \textit{The proposed $\mathcal{L}_{\text{3D}}$ enables learning from raw visual signals effectively}.
    }
    \label{tab:semi}
    \vspace{-10pt}
\end{table}

\paragraph{Training Costs.}
We analyze the extra computational costs brought by the proposed two 3D-aware visual pretext tasks in \Cref{tab:cost}.
Evaluations are conducted using 8 A100 GPUs with a global batch size of 256. 
Due to the limited GPU memory, we accumulate 32 gradient steps and the batch size per GPU is 1. 
The whole stage requires 871 training steps. GPU memories are averaged over 8 GPUs with DeepSpeed Zero 3.
As shown in the table, the denoising process introduces a \textit{negligible} increase.

\begin{table}[t]
    \centering\small
    \setlength{\tabcolsep}{11pt}
    \begin{tabular}{cc llll}
    \toprule
    $\mathcal{L}_{\text{3D}}^{\text{cross}}$ & $\mathcal{L}_{\text{3D}}^{\text{global}}$ & Speed (s/iter) & GPU Memory \\
    \midrule
    -- & -- & 111.6 & 57.2 G \\
    \rowcolor{Light}
    \checkmark & \checkmark & 125.2 (1.12$\times$) & 58.6 G (1.02$\times$) \\
    \bottomrule
    \end{tabular}
    \vspace{-5pt}
    \caption{
    \textbf{Training cost comparison.}
    All entries are based on LLaVA-Video-7B~\cite{zhang2024llavavideo} with 32 frames as inputs.
    %
    %
    $\mathcal{L}_{\text{3D}}$ brings \textit{marginal} extra computational costs.
    }
    \label{tab:cost}
\end{table}

\begin{figure}
    \centering
    \includegraphics[width=1\linewidth]{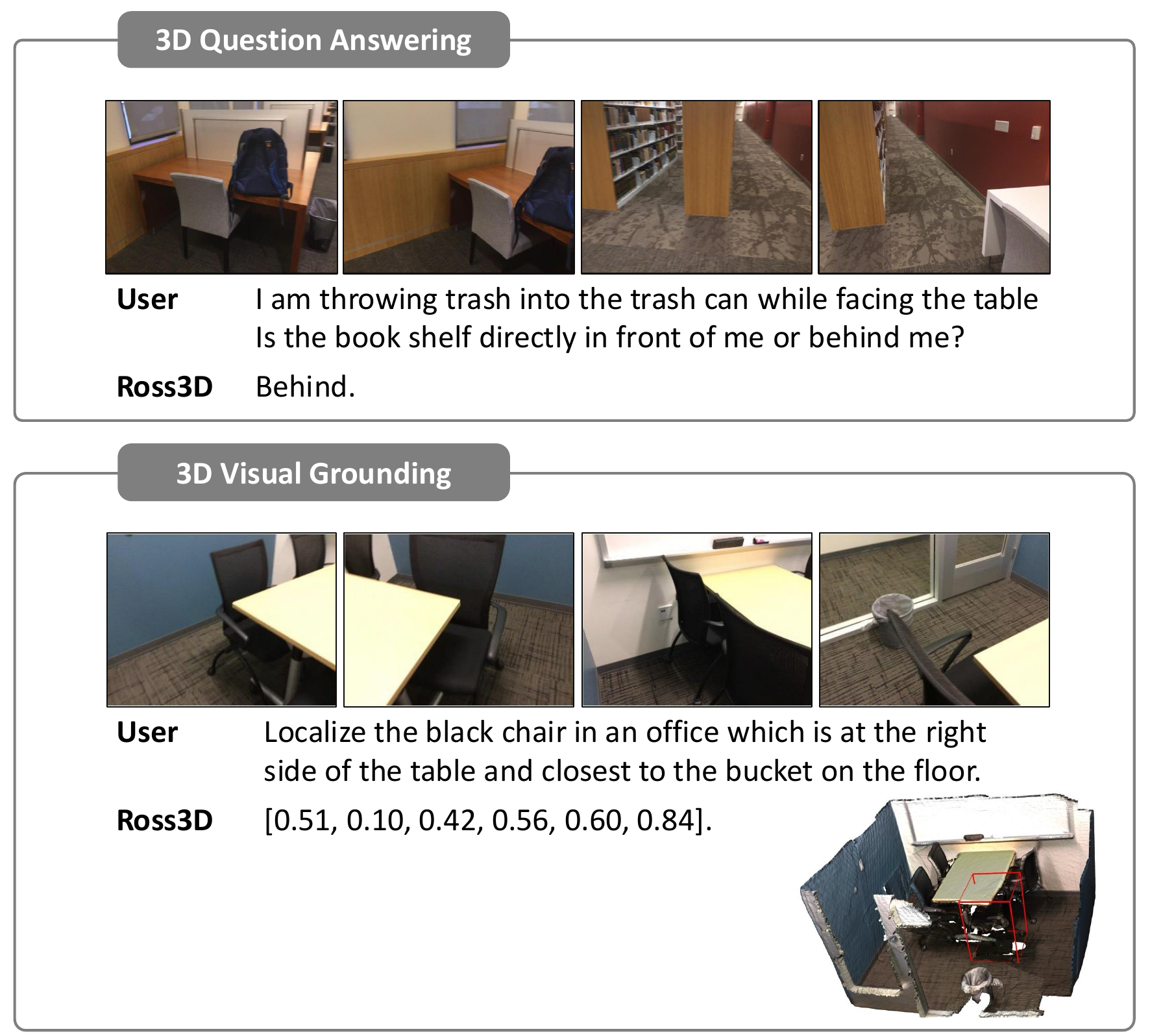}
    \vspace{-20pt}
    \caption{
    \textbf{Qualitative results.}
    Examples of 3D question answering and 3D visual grounding are sampled from ScanQA$_{\text{val}}$~\cite{ma2023sqa3d} and ScanRefer$_{\text{val}}$~\cite{chen2020scanrefer}, respectively.
    }
    \label{fig:visual}
    \vspace{-10pt}
\end{figure}

\paragraph{Qualitative Results.}
We provide qualitative results on both 3D question answering and 3D visual grounding in \Cref{fig:visual}.
%
%
Our \method is able to understand the whole scene comprehensively on 3D question answering, and perceive fine-grained details on 3D visual grounding.

\section{Conclusion}\label{sec:conclusion}

We introduce \method that significantly enhances the 3D scene understanding capabilities of LMMs.
Different from previous attempts that preliminarily focus on crafting 3D-aware input representations, we incorporate visual pretext tasks as 3D-aware supervision signals. 
These tasks, including cross-view and global-view reconstructions, enable accurate spatial relationship modeling and comprehensive scene layout comprehension, respectively.
\method demonstrates substantial improvements across various benchmarks compared to previous state-of-the-art techniques. 
More importantly, semi-supervised learning by training on 50\% text-labeled data and applying the proposed 3D visual objectives on other 50\% vision-only data even surpasses the 100\% text-supervised baseline in certain settings, demonstrating the significant potential of leveraging large amounts of unlabeled 3D data.
We hope that our research draws the community's attention to the design of 3D-aware visual supervision signals for 3D LMMs.

\section*{Acknowledgements}
The work was supported by the National Science and Technology Major Project of China (No. 2023ZD0121300), the National Natural Science Foundation of China (No. U21B2042, No. 62320106010), and the 2035 Innovation Program of CAS.

{
    \small
    \bibliographystyle{ieeenat_fullname}
    \bibliography{main}
}

\clearpage
\renewcommand\thefigure{S\arabic{figure}}
\renewcommand\thetable{S\arabic{table}}  
\renewcommand\theequation{S\arabic{equation}}
\setcounter{equation}{0}
\setcounter{table}{0}
\setcounter{figure}{0}
\setcounter{section}{0}
\renewcommand\thesection{\Alph{section}}

\section*{Supplementary Material}


\section{More Implementation Details}\label{supp:details}

\subsection{Position-Aware Video Representation}

To inject 3D information into vanilla video frames, this paper utilizes the representation proposed by~\cite{zheng2024video3dllm}.
Specifically, it adopts sinusoidal position encoding on absolute 3D coordinates $(x, y, z)$, where the coordinate of the pixel located at $(i, j)$ is computed using depth maps $\bm{D} \in \mathbb{R}^{H\times W}$, the extrinsic matrix $\bm{T} \in \mathbb{R}^{4 \times 4}$, and a camera intrinsic matrix $\bm{K} \in \mathbb{R}^{3 \times 3}$
\begin{equation}
    \begin{bmatrix}
        x & y & z & 1 
    \end{bmatrix}
    =
    \begin{bmatrix}
        \bm{D}_{ij} \cdot \begin{bmatrix}
            j & i & 1
        \end{bmatrix} \cdot (\bm{K}^{-1})^\top
         & 1
    \end{bmatrix} \cdot \bm{T}^\top.
\end{equation}
The encoded positions are then added with the original video features extracted by the vision backbone, \textit{e.g.}, CLIP~\cite{radford2021learning}.

\subsection{Training Dataset}

Our \method is a generalist model that handles multiple tasks within a single set of learned parameters.
To achieve this, \method is trained on a combined dataset, including 3D question answering dataset~\cite{ma2023sqa3d, azuma2022scanqa}, 3D dense captioning dataset~\cite{chen2021scan2cap}, and 3D visual grounding dataset~\cite{chen2020scanrefer, zhang2023multi3drefer}, in the multi-task manner similar to~\cite{zheng2024video3dllm}.

The statistics training set is illustrated in \Cref{tab:dataset}.
All data have been converted to the format of LLaVA~\cite{liu2023visual}.
There are 223K training samples in total.

\subsection{Training Objectives}

For general 3D scene understanding tasks such as 3D question answering and 3D dense captioning, we use cross-entropy loss to supervise text outputs and our proposed denoising loss to supervise visual outputs.
For 3D visual grounding, to locate more accurately, we only use 3D visual grounding loss introduced next.

We follow previous works~\cite{zheng2024video3dllm, wang2023chat, huang2022multi, zhu20233dvista} and regard the visual grounding task as a classification problem for specific object proposals.
Specifically, given a list of object proposals, we obtain object features for each object by aggregating visual embeddings. 
For each object with a bounding box $b_i$, we average the features of patches where more than 50\% of their points lie within $b_i$.
These object features are then added with the 3D position embedding of the center coordinate.
InfoNCE~\cite{oord2018representation, wang2022semi, wang2024using, wang2024pulling} is applied to optimize the similarity between the ground truth object feature and the hidden states of the special {\small\texttt{<ground>}} token.

\subsection{Evaluation Details}

For ScanRefer~\cite{chen2020scanrefer}, we simply select the object proposal with the highest similarity as the prediction.
For Multi3DRefer~\cite{zhang2023multi3drefer}, we choose the objects with the highest probabilities until the cumulative probability of selecting these objects surpasses 25\%.
For Scan2Cap~\cite{chen2021scan2cap}, we follow~\cite{zheng2024video3dllm, huang2023leo} to evaluate the captioning performance by inserting special {\texttt{<sos>}} and {\texttt{<eos>}} tokens at the start and end of the prediction, respectively. 
Greedy sampling is utilized for both 3D dense captioning and 3D question answering tasks.

\begin{table}[t]
    \centering\small
    \setlength{\tabcolsep}{4pt}
    \begin{tabular}{l cccc}
    \toprule
    Source & \# samples & \# scenes & \begin{tabular}[c]{@{}c@{}}Question\\ Length\end{tabular} & \begin{tabular}[c]{@{}c@{}}Answer\\ Length\end{tabular} \\
    \midrule
    SQA3D~\cite{ma2023sqa3d} & 79,445 & 518 & 37.8 & 1.1 \\
    ScanQA~\cite{azuma2022scanqa} & 26,515 & 562 & 13.7 & 2.4 \\
    Scan2Cap~\cite{chen2021scan2cap} & 36,665 & 562 & 13.0 & 17.9 \\
    ScanRefer~\cite{chen2020scanrefer} & 36,665 & 562 & 24.9 & -- \\
    Multi3DRefer~\cite{zhang2023multi3drefer} & 43,838 & 562 & 34.8 & -- \\
    \bottomrule
    \end{tabular}
    \vspace{-5pt}
    \caption{
    \textbf{Detailed statistics for training data.} 
    Average lengths for questions and answers are obtained from~\cite{zheng2024video3dllm}.
    }
    \label{tab:dataset}
\end{table}

\begin{table}[t]
    \centering\small
    \setlength{\tabcolsep}{6.9pt}
    \begin{tabular}{c cccc}
    \toprule
    $\gamma$ & SQA3D & ScanQA & ScanRefer & Multi3DRefer \\
    \midrule
    0.125 & 62.0 & 105.6 & 60.2 & 59.1 \\
    \rowcolor{Light}
    0.25 & \textbf{63.0} & \textbf{107.0} & \textbf{61.1} & \textbf{59.6} \\
    0.5 & 61.8 & 105.3 & 60.8 & \textbf{59.6} \\
    0.75 & 61.2 & 104.9 & 60.8 & 59.0 \\
    \bottomrule
    \end{tabular}
    \vspace{-5pt}
    \caption{
    \textbf{Ablations on the masking ratio $\gamma$.}
    A relatively small masking ratio performs slightly better, but overall, \method is robust against $\gamma$.
    }
    \label{tab:abl_cross_ratio}
\end{table}

\begin{table}[t]
    \centering\small
    \setlength{\tabcolsep}{7.8pt}
    \begin{tabular}{c cccc}
    \toprule
    $\Delta t$ & SQA3D & ScanQA & ScanRefer & Multi3DRefer \\
    \midrule
    \rowcolor{Light}
    4 & \textbf{63.0} & \textbf{107.0} & 61.1 & \textbf{59.6} \\
    2 & 62.6 & 105.4 & 60.9 & 59.2 \\
    1 & 61.8 & 104.8 & \textbf{61.2} & 59.5 \\
    \bottomrule
    \end{tabular}
    \vspace{-5pt}
    \caption{
    \textbf{Ablations on the interval $\Delta t$.}
    We implement our $\mathcal{L}_{\text{3D}}^{\text{corss}}$ and $\mathcal{L}_{\text{3D}}^{\text{global}}$ every $\Delta t$ steps.
    }
    \label{tab:abl_cross_prob}
\end{table}

\begin{table}[t]
    \centering\small
    \setlength{\tabcolsep}{8pt}
    \begin{tabular}{c cccc}
    \toprule
    BEV res. & filter & SQA3D & ScanQA & ScanRefer \\
    \midrule
    256$\times$256 & \checkmark & 62.3 & 106.5 & 60.9 \\
    432$\times$432 & -- & 61.8 & 104.6 & 60.2 \\
    \rowcolor{Light}
    432$\times$432 & \checkmark & \textbf{63.0} & \textbf{107.0} & 61.1 \\
    1024$\times$1024 & \checkmark & 62.7 & 106.5 & \textbf{61.4} \\
    \bottomrule
    \end{tabular}
    \vspace{-5pt}
    \caption{
    \textbf{Ablations on global-view reconstruction.}
    ``Filter'' indicates whether filtering out black spaces or not.
    }
    \label{tab:abl_global}
    \vspace{-10pt}
\end{table}







\section{More Experiments}\label{supp:exp}

\subsection{More Ablation Studies}

\paragraph{Design Choices for Cross-View Reconstruction.}
We ablate the masking ratio $\gamma$ and the interval $\Delta t$ in \Cref{tab:abl_cross_ratio} and \Cref{tab:abl_cross_prob}, respectively.
These designs alleviate the discrepancy between training and testing.
Empirically, a relatively \textit{small masking ratio}, \textit{i.e.}, 25\%, together with an appropriate interval, \textit{i.e.}, 4, perform the best among others.
But overall, \method is robust against these designs.

\paragraph{Design Choices for Global-View Reconstruction.}
We ablate the BEV resolution and the filtering technique in \Cref{tab:abl_global}.
\method is quite robust against these designs.

\subsection{Full Comparison}

We present full comparisons with previous approaches with the complete metrics for all benchmarks.
Specifically, we provide \Cref{tab:full_sqa} for SQA3D~\cite{ma2023sqa3d}, \Cref{tab:full_scanqa} for ScanQA~\cite{azuma2022scanqa}, \Cref{tab:full_scanrefer} for ScanRefer~\cite{chen2020scanrefer}, and \Cref{tab:full_multi3dref} for Multi3DRefer~\cite{zhang2023multi3drefer}, respectively.
Our \method significantly outperforms across all benchmarks, highlighting the effectiveness of 3D-aware visual supervision for 3D LMMs.

\begin{table*}[t]
    \centering\small
    \setlength{\tabcolsep}{12.7pt}
    \begin{tabular}{l cccccc cc}
    \toprule
    \multirow{2}{*}{Method} & \multicolumn{6}{c}{Question Type} & \multirow{2}{*}{Avg. (EM)} & \multirow{2}{*}{EM-R} \\
    \cmidrule(lr){2-7}
    & What & Is & How & Can & Which & Others \\
    \midrule
    \multicolumn{9}{l}{\textit{Expert Models}} \\
    SQA3D~\cite{ma2023sqa3d} & 31.6 & 63.8 & 46.0 & 69.5 & 43.9 & 45.3 & 46.6 & -- \\
    3D-VisTA~\cite{zhu20233dvista} & 34.8 & 63.3 & 45.4 & 69.8 & 47.2 & 48.1 & 48.5 & -- \\
    \midrule
    \multicolumn{9}{l}{\textit{2D LLMs}} \\
    InternVL2-8B~\cite{internvl2} & 30.5 & 53.8 & 5.5 & 47.3 & 25.8 & 36.3 & 33.0 & 45.3 \\
    Qwen2-VL-7B~\cite{wang2024qwen2vl} & 29.0 & 59.2 & 33.4 & 50.5 & 44.2 & 43.2 & 40.7 & 46.7 \\
    LLaVA-Video-7B~\cite{zhang2024llavavideo} & 42.7 & 56.3 & 47.5 & 55.3 & 50.1 & 47.2 & 48.5 & -- \\
    \midrule
    \multicolumn{9}{l}{\textit{3D LMMs}} \\
    LEO~\cite{huang2023leo} & -- & -- & -- & -- & -- & -- & 50.0 & 52.4 \\
    Scene-LLM~\cite{fu2024scene} & 40.9 & 69.1 & 45.0 & \textbf{70.8} & 47.2 & 52.3 & 54.2 & -- \\
    ChatScene~\cite{zhang2024chatscene} & 45.4 & 67.0 & 52.0 & 69.5 & 49.9 & 55.0 & 54.6 & 57.5 \\
    LLaVA-3D~\cite{zhu2024llava3d} & -- & -- & -- & -- & -- & -- & 55.6 & -- \\
    Video-3D-LLM~\cite{zheng2024video3dllm} & 51.1 & 72.4 & 55.5 & 69.8 & 51.3 & 56.0 & 58.6 & -- \\
    \textcolor{codegray}{GPT4Scene-HDM$^{\ddag}$}~\cite{qi2025gpt4scene} & \textcolor{codegray}{55.9} & \textcolor{codegray}{69.9} & \textcolor{codegray}{50.8} & \textcolor{codegray}{68.7} & \textcolor{codegray}{53.3} & \textcolor{codegray}{60.4} & \textcolor{codegray}{59.4} & \textcolor{codegray}{62.4} \\
    \rowcolor{Light}
    \method & \textbf{56.0} & \textbf{79.8} & \textbf{60.6} & 70.4 & \textbf{55.3} & \textbf{60.1} & \textbf{63.0} & \textbf{65.7} \\
    \bottomrule
    \end{tabular}
    \vspace{-5pt}
    \caption{
    \textbf{Full comparison of 3D question answering} on SQA3D~\cite{ma2023sqa3d} test set.
    ``$\ddag$'' indicates this result is achieved by adopting a larger input resolution (512$\times$490) and incorporating extra BEV inputs.
    }
    \label{tab:full_sqa}
\end{table*}

\begin{table*}[t]
    \centering\small
    \setlength{\tabcolsep}{9.8pt}
    \begin{tabular}{l c cccc ccc}
    \toprule
    \multirow{2}{*}{Method} & & \multicolumn{4}{c}{BLEU-n Metrics} & \multicolumn{3}{c}{Language Generation Metrics} \\
    \cmidrule(lr){3-6}
    \cmidrule(lr){7-9}
    & EM & BLEU-1 & BLEU-2 & BLEU-3 & BLEU-4 & ROUGE & METEOR & CIDEr \\
    \midrule
    \multicolumn{9}{l}{\textit{Expert Models}} \\
    ScanQA~\cite{azuma2022scanqa} & 21.1 & 30.2 & 20.4 & 15.1 & 10.1 & 33.3 & 13.1 & 64.9 \\
    3D-VLP~\cite{jin2023context} & 21.7 & 30.5 & 21.3 & 16.7 & 11.2 & 34.5 & 13.5 & 67.0 \\
    3D-VisTA~\cite{zhu20233dvista} & -- & -- & -- & -- & 13.9 & 35.7 & -- & -- \\
    \midrule
    \multicolumn{9}{l}{\textit{2D LLMs}} \\
    InternVL2-8B~\cite{internvl2} & 16.9 & 20.0 & 9.8 & 5.2 & 2.7 & 32.6 & 14.5 & 55.3 \\
    Qwen2-VL-7B~\cite{wang2024qwen2vl} & 19.0 & 27.8 & 13.6 & 6.3 & 3.0 & 34.2 & 11.4 & 53.9 \\
    LLaVA-Video-7B~\cite{zhang2024llavavideo} & -- & 39.7 & 26.6 & 9.3 & 3.1 & 44.6 & 17.7 & 88.7 \\
    \midrule
    \multicolumn{9}{l}{\textit{3D LMMs}} \\
    3D-LLM~\cite{hong20233d} & 20.5 & 39.3 & 25.2 & 18.4 & 12.0 & 35.7 & 14.5 & 69.4 \\
    Chat-3D~\cite{wang2023chat} & -- & 29.1 & -- & -- & 6.4 & 28.5 & 11.9 & 53.2 \\
    LL3DA~\cite{chen2024ll3da} & -- & -- & -- & -- & 13.5 & 37.3 & 15.9 & 76.8 \\
    LEO~\cite{huang2023leo} & 24.5 & -- & -- & -- & 11.5 & 39.3 & 16.2 & 80.0 \\
    Scene-LLM~\cite{fu2024scene} & 27.2 & 43.6 & 26.8 & 19.1 & 12.0 & 40.0 & 16.6 & 80.0 \\
    ChatScene~\cite{zhang2024chatscene} & 21.6 & 43.2 & 29.1 & 20.6 & 14.3 & 41.6 & 18.0 & 87.7 \\
    Grounded 3D-LLM~\cite{chen2024grounded} & -- & -- & -- & -- & 13.4 & -- & -- & 72.7 \\
    LLaVA-3D~\cite{zhu2024llava3d} & 27.0 & -- & -- & -- & 14.5 & 50.1 & 20.7 & 91.7 \\
    Video-3D-LLM~\cite{zheng2024video3dllm} & 30.1 & 47.1 & 31.7 & 22.8 & 16.2 & 49.0 & 19.8 & 102.1 \\
    \textcolor{codegray}{GPT4Scene-HDM$^{\ddag}$}~\cite{qi2025gpt4scene} & \textcolor{codegray}{28.2} & \textcolor{codegray}{44.4} & \textcolor{codegray}{30.3} & \textcolor{codegray}{22.3} & \textcolor{codegray}{15.5} & \textcolor{codegray}{46.5} & \textcolor{codegray}{18.9} & \textcolor{codegray}{96.3} \\
    \rowcolor{Light}
    \method & \textbf{30.8} & \textbf{49.2} & \textbf{33.7} & \textbf{24.9} & \textbf{17.9} & \textbf{50.7} & \textbf{20.9} & \textbf{107.0} \\
    \bottomrule
    \end{tabular}
    \vspace{-5pt}
    \caption{
    \textbf{Full comparison of 3D question answering} on ScanQA~\cite{ma2023sqa3d} validation set.
    ``$\ddag$'' indicates this result is achieved by adopting a larger input resolution (512$\times$490) and incorporating extra BEV inputs.
    }
    \label{tab:full_scanqa}
\end{table*}

\begin{table*}[t]
    \centering\small
    \setlength{\tabcolsep}{13.5pt}
    \begin{tabular}{l cc cc cc}
    \toprule
    \multirow{2}{*}{Method} & \multicolumn{2}{c}{Unique} & \multicolumn{2}{c}{Multiple} & \multicolumn{2}{c}{Overall} \\
    \cmidrule(lr){2-3}
    \cmidrule(lr){4-5}
    \cmidrule(lr){6-7}
    & Acc@0.25 & Acc@0.5 & Acc@0.25 & Acc@0.5 & Acc@0.25 & Acc@0.5 \\
    \midrule
    \multicolumn{7}{l}{\textit{Expert Models}} \\
    ScanRefer~\cite{chen2020scanrefer} & 76.3 & 53.5 & 32.7 & 21.1 & 41.2 & 27.4 \\
    3D-VLP~\cite{jin2023context} & 84.2 & 64.6 & 43.5 & 33.4 & 51.4 & 39.5 \\
    3D-VisTA~\cite{zhu20233dvista} & 81.6 & 75.1 & 43.7 & 39.1 & 50.6 & 45.8 \\
    MVT~\cite{huang2022multi} & 77.7 & 66.5 & 31.9 & 25.3 & 40.8 & 33.3 \\
    3DVG-Trans~\cite{zhao20213dvg} & 81.9 & 60.6 & 39.3 & 28.4 & 47.6 & 34.7 \\
    ViL3DRel~\cite{chen2022language} & 81.6 & 68.6 & 40.3 & 30.7 & 47.9 & 37.7 \\
    3DJCG~\cite{cai20223djcg} & 83.4 & 64.3 & 41.4 & 30.8 & 49.6 & 37.3 \\
    M3DRef-CLIP~\cite{zhang2023multi3drefer} & 85.3 & 77.2 & 43.8 & 36.8 & 51.9 & 44.7 \\
    \midrule
    \multicolumn{7}{l}{\textit{3D LMMs}} \\
    3D-LLM~\cite{hong20233d} & -- & -- & -- & -- & 30.3 & -- \\
    Grounded 3D-LLM~\cite{chen2024grounded} & -- & -- & -- & -- & 47.9 & 44.1 \\
    LLaVA-3D~\cite{zhu2024llava3d} & -- & -- & -- & -- & 54.1 & 42.2 \\
    ChatScene~\cite{zhang2024chatscene} & \textbf{89.6} & \textbf{82.5} & 47.8 & 42.9 & 55.5 & 50.2 \\
    Video-3D-LLM~\cite{zheng2024video3dllm} & 88.0 & 78.3 & 50.9 & 45.3 & 58.1 & 51.7 \\
    \textcolor{codegray}{GPT4Scene-HDM$^{\ddag}$}~\cite{qi2025gpt4scene} & \textcolor{codegray}{90.3} & \textcolor{codegray}{83.7} & \textcolor{codegray}{56.4} & \textcolor{codegray}{50.9} & \textcolor{codegray}{62.6} & \textcolor{codegray}{57.0} \\
    \rowcolor{Light}
    \method & 87.2 & 77.4 & \textbf{54.8} & \textbf{48.9} & \textbf{61.1} & \textbf{54.4} \\
    \bottomrule
    \end{tabular}
    \vspace{-5pt}
    \caption{
    \textbf{Full comparison of 3D visual grouding} on ScanRefer~\cite{chen2020scanrefer} validation set.
    ``$\ddag$'' indicates this result is achieved by adopting a larger input resolution (512$\times$490) and incorporating extra BEV inputs.
    ``Unique'' and ``Multiple'' depend on whether there are other objects of the same class as the target object.
    }
    \label{tab:full_scanrefer}
\end{table*}

\begin{table*}[t]
    \centering\small
    \setlength{\tabcolsep}{4.1pt}
    \begin{tabular}{l c c cc cc cc cc}
    \toprule
    \multirow{2}{*}{Method} & ZT w/o D & ZT w/ D & \multicolumn{2}{c}{ST w/o D} & \multicolumn{2}{c}{ST w/ D} & \multicolumn{2}{c}{MT} & \multicolumn{2}{c}{ALL}\\
    \cmidrule(lr){2-2}
    \cmidrule(lr){3-3}
    \cmidrule(lr){4-5}
    \cmidrule(lr){6-7}
    \cmidrule(lr){8-9}
    \cmidrule(lr){10-11}
    & F1 & F1 & F1@0.25 & F1@0.5 & F1@0.25 & F1@0.5 & F1@0.25 & F1@0.5 & F1@0.25 & F1@0.5 \\
    \midrule
    \multicolumn{7}{l}{\textit{Expert Models}} \\
    3DVG-Trans~\cite{zhao20213dvg} & 87.1 & 45.8 & -- & 27.5 & -- & 16.7 & -- & 26.5 & -- & 25.5 \\
    M3DRef-CLIP~\cite{zhang2023multi3drefer} & 81.8 & 39.4 & 53.5 & 47.8 & 34.6 & 30.6 & 43.6 & 37.9 & 42.8 & 38.4 \\
    3DJCG~\cite{cai20223djcg} & 94.1 & 66.9 & -- & 26.0 & -- & 16.7 & -- & 26.2 & -- & 26.6 \\
    \midrule
    \multicolumn{7}{l}{\textit{3D LMMs}} \\
    ChatScene~\cite{zhang2024chatscene} & 90.3 & 62.6 & \textbf{82.9} & \textbf{75.9} & 49.1 & 44.5 & \textbf{45.7} & \textbf{41.1} & 57.1 & 52.4 \\
    Video-3D-LLM~\cite{zheng2024video3dllm} & \textbf{94.7} & \textbf{78.5} & 82.6 & 73.4 & 52.1 & 47.2 & 40.8 & 35.7 & 58.0 & 52.7 \\
    \textcolor{codegray}{GPT4Scene-HDM$^{\ddag}$}~\cite{qi2025gpt4scene} & \textcolor{codegray}{97.4} & \textcolor{codegray}{84.4} & \textcolor{codegray}{85.0} & \textcolor{codegray}{77.7} & \textcolor{codegray}{59.9} & \textcolor{codegray}{55.1} & \textcolor{codegray}{48.6} & \textcolor{codegray}{44.6} & \textcolor{codegray}{64.5} & \textcolor{codegray}{59.8} \\
    \rowcolor{Light}
    \method & 93.6 & 77.8 & 80.2 & 72.1 & \textbf{54.7} & \textbf{49.6} & 44.3 & 39.1 & \textbf{59.6} & \textbf{54.3} \\
    \bottomrule
    \end{tabular}
    \vspace{-5pt}
    \caption{
    \textbf{Full comparison of 3D visual grouding} on Multi3DRefer~\cite{zhang2023multi3drefer} validation set.
    ``$\ddag$'' indicates this result is achieved by adopting a larger input resolution (512$\times$490) and incorporating extra BEV inputs.
    ``ZT'' means zero-target.
    ``ST'' denotes single-target and ``MT'' is multi-target.
    ``D'' indicates distractor.
    }
    \label{tab:full_multi3dref}
\end{table*}


\end{document}